\documentclass[letterpaper, 10 pt, conference]{ieeeconf}  

\IEEEoverridecommandlockouts                              

\usepackage{graphics} 
\usepackage[pdftex]{graphicx}
\usepackage{multirow}
\usepackage{slashbox}

\title{\LARGE \bf
Measuring the Quality of Exercises
}

\author{Paritosh Parmar$^{1}$ and Brendan Tran Morris$^{2}$
\thanks{$^{1}$Real-Time Intelligent Systems (RTIS) Laboratory, University of Nevada, Las Vegas
        {\tt\small parmap1@unlv.nevada.edu}}%
\thanks{$^{2}$Faculty of Electrical and Computer Engineering, University of Nevada, Las Vegas,
        {\tt\small brendan.morris@unlv.edu}}%
}

\begin{document}

\maketitle
\thispagestyle{empty}
\pagestyle{empty}

\begin{abstract}

This work explores the problem of exercise quality measurement since it is essential for effective management of diseases like cerebral palsy (CP).  This work examines the assessment of quality of large amplitude movement (LAM) exercises designed to treat CP in an automated fashion.  Exercise data was collected by trained participants to generate ideal examples to use as a positive samples for machine learning.  Following that, subjects were asked to deliberately make subtle errors during the exercise, such as restricting movements, as is commonly seen in cases of patients suffering from CP.  The quality measurement problem was then posed as a classification to determine whether an example exercise was either "good" or "bad". Popular machine learning techniques for classification, including support vector machines (SVM), single and double-layered neural networks (NN), boosted decision trees, and dynamic time warping (DTW), were compared. The AdaBoosted tree performed best with an accuracy of 94.68\% demonstrating the feasibility of assessing exercise quality.

\end{abstract}

\section{Introduction}

Currently, there are no available cures for mobility disease like CP.  However, these disease can be managed, i.e., muscle function and coordination can be improved with appropriate intervention.  Review of intervention literature shows that physical therapy (PT) at home is one of the best methods to effectively manage CP \cite{novak}.  This type of PT session employs physical therapists who are able to effectively judge the quality of an exercise performed by a patient to design an effective personalized intervention program.  

However, a major factor for achieving improvement is a high volume of the right exercises \cite{chang}.  This means a patient should have as many sessions with a physical therapist as possible which in turn may cause a financial burden.  It can be found in the work by Wu et al. \cite{wu} that children of socioeconomically disadvantaged populations are at high risk or suffering from CP.  If the cost of treatment and PT sessions could be reduced, more diverse populations would be able to afford and benefit from CP management techniques.  

This work suggests an automated system to assess exercise quality to complement traditional PT.  A camera-based system using the commercial Microsoft Kinect sensor is used to observe patients at home while performing exercises.  This supplements physical therapist visits to increase exercise volume.  This work presents a comparison of various machine learning techniques to automatically assess the quality of an exercise and to determine if it was accomplished satisfactorily.  The exercises of interest are LAM which were specifically designed for children with CP by PT \cite{hickman}.  The experiments show that it is possible to automatically and effectively recognize a high-quality versus a low-quality repetition of an exercise as demonstrated in Fig. \ref{fig:data_samples}.

The paper is organized as follows: Related works are discussed first.  The following section describes a new LAM Exercise Quality Dataset that was collected and used in machine learning techniques used.  In Section \ref{sec:methods}, different methods to assess the quality of exercises are presented.  Finally, after presenting the experimental results in Section V, we conclude our work and discuss future directions in Section \ref{sec:conclusion}.

\begin{figure}[t!]
\centering
\includegraphics[width=\columnwidth, height=6cm]{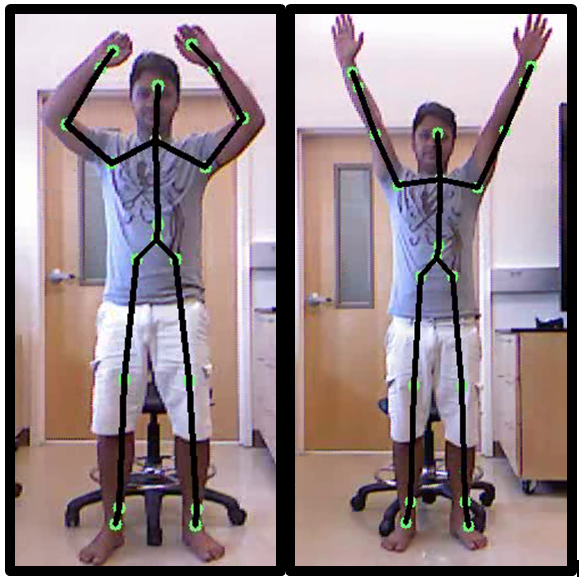}
\caption{Blastoff Exercise:  The left image is a "bad" exercise repetition with error in LAM due to restricited arm extension.  The right image is the ideal or "good" exercise repetition.}
\label{fig:data_samples}
\end{figure}

\section{Related Work}

Assessing quality of exercises (equivalently, actions) while seemingly similar to, is quite different from action recognition. A supervised action recognition system is trained for certain specific actions, while quality can have any value between the upper and lower limits for any action.  The second distinction is that, generally, there would be significant amount of difference between two classes of actions, while, in case of quality assessment, we need to look out for very subtle differences (errors). In this section, we highlight works that are directly related to quality assessment of actions.

Gordon discusses three important issues in automated assessment of human performance \cite{gordon95}. The three issues discussed are: appropriate performance types, computer vision technology required for assessing and effects of automation on performance assessment concerns. He concludes that tracking was best suited for this task and that features or parameters used for assessment should be observable from the video or could be inferred from video. 

Jug et al., discuss an approach to assess coordinated human activity on the basis of trajectories \cite{jug}. They considered basketball and developed a probabilistic method to detect key events.

The FINA09 diving dataset was introduced in \cite{wnuk}. They carry out an initial investigation and conclude that temporal information, implicitly present in videos, is very important and that HOG features can be used for pose estimation. Pirsiavash et al., addressed the issue of quality assessment \cite{quality}. They developed an approach in which the system is trained on an action dataset which has scores, for quality, marked by humans. Once trained, the system can assess sample and output a score for quality between 0 and 100. They tested their system on Olympic diving dataset in which the scores were given by judges.

While there have been some limited work in the area of quality assessment of actions, there is much that remains to be explored.

\section{LAM Exercise Quality Dataset}

Before explaining the methods to assess exercise quality, we first discuss the exercise dataset.  The motivational goal is to have an automated system that can tell whether a patient made an error during a repetition of an exercise.  With this ability, a patient could perform home therapy and still get constructive feedback.  

The dataset contains five LAM exercises \{Blast-Off, Body-Builder, Finish-Line, Reach-For-The-Stars, Take-A-Bow\} performed by five different male subjects with differing heights and ages.  Each subject was trained and instructed to perform the exercise as precisely as possible.  These examples represented the positive or "good" samples since these exercises do not contain significant errors.  The exercises were then repeated only now with significant errors introduced -- such as by restricting movements as shown in Fig. \ref{fig:data_samples}.  

\begin{table}[]
\centering
\caption{Details of Dataset for Blast-Off Exercise}
\label{my-label}
\begin{tabular}{|r|c|c|c|c|c|c|}
\hline
\textbf{Subject}          & \textbf{1} & \textbf{2} & \textbf{3} & \textbf{4} & \textbf{5} & \textbf{Total} \\ \hline
\textbf{Negative Samples} & 10         & 13         & 10         & 14         & 15         & 62             \\ \hline
\textbf{Positive Samples} & 11         & 13         & 10         & 14         & 15         & 63             \\ \hline
\end{tabular}
\end{table}

\begin{figure}[tbh]
\centering
\includegraphics[width=\columnwidth,height=4cm]{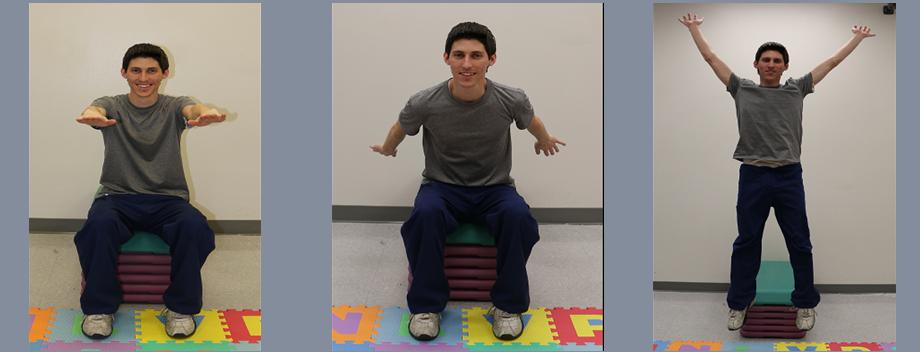}
\caption{A physical therapy expert demonstrating the three steps to perform the Blastoff exercise: i) sitting with arms stretched in front, ii) arms swung back in behind stretch position, and iii) explosively standing up with arms full extended up.}
\label{fig:blastoff}
\end{figure}

The subjects used to generate the LAM Exercise Dataset were not actual CP patients but healthy individuals.  Each exercise was formally defined and the subjects received instructions on how to perform exercises with images and training videos from physical therapist that were experts in the LAM protocols.  In particular, each exercise was broken into important phases as shown in Fig. \ref{fig:blastoff}.  With the "good" samples, participants focuses on these critical steps of the exercise to ensure there were no mistakes during these phases.  In addition, the subjects were observed during test trials before data collection to improve performance.

The negative samples were collected after subjects were instructed to make errors during an exercise to simulate mobility problems seen in CP patients.  The errors were designed to be functionally significant in performance but not enough to lose the overall goal and spirit of a LAM exercise.  The deliberate errors could sometimes be subtle and included mistakes such as not stretching an arm completely or not completely transitioning to a critical position between exercise phases.  

The data collection was performed using a Microsoft Kinect sensor which provides color (RGB) and depth (D) information.  In addition to the RGBD data, the Kinect also provides skeleton tracking for 20 joints (e.g. shoulder, elbow, hip, knee, etc.).  The joints of interest can be seen as green dots in Fig. \ref{fig:data_samples}.  The full dataset consists of the skeleton joint positions as well as recorded video of the participant for visual inspection.  On an average, 11 positive and 11 negative samples were collected per subject for each of the exercises. Details for Blast-Off exercise samples are provided in Table I.   

\section{Methods}
\label{sec:methods}

In order to assess the quality of an exercise, various classification algorithms are compared using different methods to represent the tracked skeleton during a sample.   

\subsection{Data Preprocessing for Equal Length}

The raw data consists of 20 joint positions for each frame; an exercise sample (positive or negative) contains many such frames. However, due to differences in speed and execution, the samples do not have the same number of frames.  The samples are re-sampled to a fixed length of 160 frames for simplicity.  A feature vector can then be built from the 3D (X,Y,Z) joint positions of each frame concatenated together which has dimension $3\times20\times160=9600$. 

\subsubsection{Height Scaling}
Since people have different heights, the skeleton positions need to be normalized to a prototypical range for comparison.  Without this scaling, two examples could be similar in exercise quality but seem very different just by joint position since the length of body parts is in proportion to the height.  In this work, we scale the joint positions (X,Y,Z) between 1 and 3 (this choice is arbitrary).  Once scaled, we find relative joints positions as discussed in next subsection. 

\subsubsection{Relative Joint Positions}

Obviously, when performing an exercise, the participant will not be in the same position in front of the Kinect each time.  Since the offset from the center of the camera will not be the same, the position of the hip center is subtracted from all the joints to obtain a skeleton centric coordinate frame. In effect, this gives us positions of every joint relative to the hip center joint.

We found that first scaling and then finding relative positions gave better results than going the other way around. 

\subsubsection{Skeleton Representations}
By transforming the raw skeleton 3D time-series position data, different exercise characteristics can be examined.  Besides position, important joint angles: \{angles formed at knee, at elbow, between femur and spine, between elbow, shoulder and hip center, and between elbow, shoulder and other shoulder\} are computed from the skeleton to explicitly consider exercise extension.  Further, both joint positions and angles are transformed from the time-domain into the frequency domain by taking the discrete cosine transform (DCT). Each data variation can then be used as the input feature vector for every classifier. 

\subsection{Data Classification}
After preparing the skeleton data, the quality assessment task is treated as a binary classification problem.  Machine learning techniques are utilized to distinguish between the ideal and erroneous versions of exercises.  The classifiers considered are SVM's, NN's, boosted trees, and DTW.  

\subsection{Support Vector Machines}

The first classifier that we tested was a binary SVM since it is one of the most popular binary classifiers due to its ability to discriminate between classes with favorable generalization properties. In case of binary SVM, good exercise repetitions belong to one class, while bad exercise repetitions belong to the other class. The 9600 dimension feature vector is directly used as input to the SVM.
Apart from binary SVM, a one-class SVM was trained as well.  The one-class SVM only uses the positive training samples to build a model of the right way to do an exercise.  The one-class SVM model learns the boundary around the positive training data.  During test, any example that deviates from this ideal way is considered an anomaly or a "bad" sample.

\subsection{Boosted tree}

We tried following boosting techniques: adaboost, gentleboost, logitboost, robustboost and totalboost. Boosting is an iterative algorithm to build more complex and accurate classifiers using simple weak learners.  The weak learner utilized in this work was a tree classifier.  

\subsection{DTW}
DTW is a sequence alignment technique introduced for speech recognition.  It provides a simple mechanism to compare data of variable length.  In this work, it is used to find the closeness of an exercise repetition to its ideal form.  The ideal form was constructed as the average of all positive training samples.  A threshold is then learned with the training set to set the distance which differentiates between "good" and "bad" examples.  The learned threshold is the maximum training distance for a "good" example and hence any test example with greater distance is considered "bad".

\subsection{Neural Networks} 

Generally, a NN has an input layer some number of hidden layers and an output layer. Shallow NN have a single hidden layer while deep have many more hidden layers.

The preprocessed skeleton data is fed directly to the NN with 9600 input neurons.  Various number of neurons were tested in the shallow hidden layer with 500 neurons working reasonably well.

\subsubsection*{Multi-layer neural network}

MNN has more than one hidden layer where each layer learns a higher level representation of the input data with more parameters. MNN are known to suffer from the problem of vanishing or exploding gradient during training phase which makes them sometimes difficult to train, but tend to perform better provided that enough data is provided. 

\section{Experiments \& Results}

We tried two approaches, on the Blast-Off exercise, to check the efficiency of the classifiers discussed in the previous section. The difference between the approaches was in the training \& testing sets.

In the first approach (3 vs. 2 Training), we trained the classifiers on the data of 3 human subjects. These classifiers were then tested on the data of the remaining two human subjects. In this way, we had 78 training samples and 47 testing samples.

In the second approach (Random Sample Training), we took the samples belonging to all the human subjects and then randomly selected 80 samples from the sample pool as our training set and the remaining 45 samples as our test set. We define accuracy as the ratio of correctly classified samples to the total test samples, true positive rate as the ratio of true positive to total no. of positives, and false positive rate as the ratio of the false positives to the total no. of negatives. The results from both approaches are as follows.

\subsection{3 vs. 2 Training}

\subsubsection{SVM}
The performance of the SVM, for quality assessment was worse than chance (50\%). The reason for this extremely poor performance is due to the limited training dataset of 78 samples from only 3 human subjects.  These samples are too few to generalize well in the large 9600 dimensional feature space.  Instead of learning a robust classifier between "good" and "bad" exercises, the training data was overfit, leading to poor test data performance.
The performance of the one-class SVM was worse than the binary SVM presumably since only 39 positive samples were used during training.

\subsubsection{Boosted Trees}
Among all the boosting techniques, adaboost yielded the best results after 90 iterations.  

\subsubsection{NN}
Single-Layer NN (SNN) performed better than MNN for all the skeleton representations. 

Since our network is a fully connected one, we have more than 4.8 million parameters (weights and biases). Training any NN, is in essence, setting all these parameters. Our small training set was not enough to set 4.8 million parameters. So, NN is bound to perform poorly on the test set again due to overfitting as in the SVM case.  

The results obtained with this training are disappointing since our target PT application cannot really tolerate many misclassifications. The results of all the classifiers are summarized in Table II.

\begin{table}[]
\centering
\caption{Quality Classifier \% Accuracy (3 vs. 2 Training)}
\label{tab:accuracy_table_1}
\resizebox{\columnwidth}{!}{%
\begin{tabular}{|r||c|c|c|c|}
\hline
\multicolumn{1}{l}{\multirow{2}{*}{}} & \multicolumn{2}{c}{\textbf{Time domain}}  & \multicolumn{2}{c}{\textbf{Frequency domain}} \\
\hline
\multicolumn{1}{l}{}                  & \textbf{Joint data} & \textbf{Angle data} & \textbf{Joint data}   & \textbf{Angle data}   \\
\hline \hline
\textbf{SVM}                          & 46.81               & 51.06               & 46.81                 & 51.06                 \\
\hline
\textbf{Single-layer NN}              & 74.47               & 74.47               & \textbf{59.58}                 & \textbf{57.45} \\
\hline
\textbf{Multi-layer NN}               & 54.68               & 59.71               & 51.06                 & 42.55                     \\
\hline
\textbf{AdaBoosted tree}              & \textbf{85.11}               & 61.70               & 48.94                 & 46.81                 \\
\hline
\textbf{DTW}                          & 46.81               & \textbf{76.92}               & -                     & -                   \\ 
\hline 
\end{tabular}
}
\end{table}

\subsection{Random Sample Training}
Due to random nature of this approach, we carried out 51 runs of this experiment for each classifiers and each type data (time domain, frequency domain, angle, etc.) and averaged the results, which are shown in Table III. For the Receiver Operating Characteristic (ROC), shown in Fig. 3, we used the median of the results sorted by accuracy.

In this case, for boosted trees, we found that using 300 rounds, instead of 90, as in previous approach, of boosting yielded better results.

\begin{table}[]
\centering
\caption{Quality Classifier \% Accuracy (Random Sample Training)}
\label{tab:accuracy_table_1}
\resizebox{\columnwidth}{!}{%
\begin{tabular}{|r||c|c|c|c|}
\hline
\multicolumn{1}{l}{\multirow{2}{*}{}} & \multicolumn{2}{c}{\textbf{Time domain}}  & \multicolumn{2}{c}{\textbf{Frequency domain}} \\
\hline
\multicolumn{1}{l}{}                  & \textbf{Joint data} & \textbf{Angle data} & \textbf{Joint data}   & \textbf{Angle data}   \\
\hline \hline
\textbf{SVM}                          & 90.13               & 87.41               & \textbf{90.89 }                 & \textbf{87.63}                 \\
\hline
\textbf{Single-layer NN}              & 90.65               & 79.09               & 56.60                 & 55.96                \\
\hline
\textbf{Multi-layer NN}               & 80.21               & 74.44               & 78.35                 & 68.89                \\
\hline
\textbf{AdaBoosted tree}              & \textbf{94.68}               & \textbf{90.30 }              & 83.14                 & 80.40                 \\
\hline
\textbf{DTW}                          & 73.64               & 74.03               & -                     & -                   \\ 
\hline 
\end{tabular}
}
\end{table}

\begin{figure}[t!]
\centering
\includegraphics[width=8cm, height=4cm]{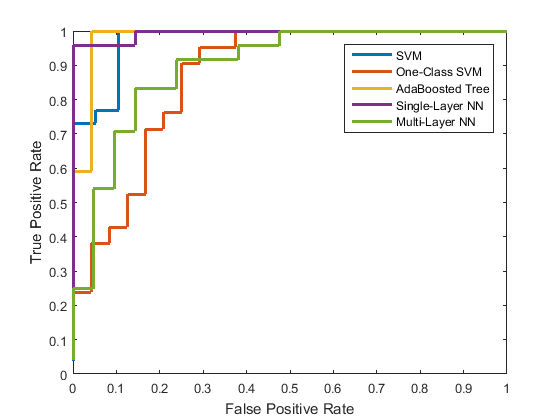}
\caption{Receiver Operating Characteristics of various classifiers.}
\label{fig:data_samples}
\end{figure}

By comparing the results obtained by following both the approaches, we can see that we got much better results in second approach. Results in case of some of the classifiers were better by more than 30\% because the classifiers are likely to have been trained on least some samples of every subject. This indicates that the classifiers were not generalizing well across the patients.

In general, we can see that we got better results with joint data in time domain than any other type of data representation. Best results were achieved by AdaBoosted trees, followed by SVM in frequency domain.

With this pilot study, machine learning techniques can be seen to have promise as an automated quality assessment tool that could potentially be used to supply immediate feedback to a patient during in-home rehabilitation.

\section{Concluding Remarks}
\label{sec:conclusion}
In this paper, the problem of exercise quality assessment was addressed using computer vision. The pilot LAM Exercise Quality Dataset was introduced to distinguish between exercises that were well performed against those that were performed poorly.  A Microsoft Kinect sensor was used to collect skeleton data which we showed needed to be pre-processed before use.  Various machine learning classifiers, such as SVM, SNN, MNN, boosted trees, and DTW, were compared on different skeleton representations to determine the most effective method to assess quality. Experiments showed that AdaBoosted tree on joint data yielded the best results with an accuracy of 94.68\%, followed by  followed by SVM in frequency domain with an accuracy of 90.89\%. 

The main limitation of this pilot study was the limited size of the training and testing datasets.  In the future, a larger study will be performed with more varied participants.  Further, rather than a binary classification, the problem will be viewed as a regression problem to determine how "good" or "bad" an exercise sample is.  Finally, when there are mistakes, a system will be developed to pinpoint in which joints more emphasis is required for improvement. 

\addtolength{\textheight}{-12cm}   





\bibliographystyle{IEEETran}
\bibliography{refs}

\end{document}